%% file: main.tex
\documentclass[letterpaper, 10 pt, conference]{ieeeconf}

\IEEEoverridecommandlockouts
\usepackage{cite}
\usepackage{amsmath,amssymb,amsfonts}
\usepackage{algorithmic}
\usepackage{graphicx}
\usepackage{textcomp}
\usepackage{xcolor}

\usepackage[table]{xcolor}

\usepackage{graphicx} 
\usepackage{tikz} 
\usepackage{booktabs}
\usepackage{comment}
\usetikzlibrary{fit, backgrounds, positioning, arrows.meta, calc, shapes.geometric}
\def\BibTeX{{\rm B\kern-.05em{\sc i\kern-.025em b}\kern-.08em
    T\kern-.1667em\lower.7ex\hbox{E}\kern-.125emX}}

\begin{document}

\title{Lang3DSeg: Annotation-Free Open-Vocabulary \\ 3D Segmentation with Point Transformers\\
}

\author{Cigdem Kokenoz$^{1}$, Amir Salarpour$^{2}$, Alkim Domeke$^{2}$, \\
Christopher Salas$^{2}$, Pedram MohajerAnsari$^{2}$, Long Cheng$^{2}$, \\ Mert D. Pesé$^{2}$,and Bing Li$^{1}$%
    \thanks{DISTRIBUTION STATEMENT A. Approved for public release; distribution is unlimited. [OPSEC 11142].}%
    \thanks{$^{1}$Department of Automotive Engineering, Clemson University, Greenville, SC, USA. Corresponding author: {\tt\small ckokeno@clemson.edu}.}%
    \thanks{$^{2}$School of Computing, Clemson University, Clemson, SC, USA.}%
    \thanks{Submitted to the IEEE International Conference on Robotics and Automation (ICRA) 2027. Code will be released upon acceptance at {\tt\small https://github.com/<org>/Lang3DSeg}.}%
}

\maketitle

\input{contents/0-abstract}

\section{Introduction}
\input{contents/1-introduction}

\section{Related Work}
\input{contents/2-related_work}

\input{figures/framework}

\section{Methodology}
\input{contents/3-methods}

\section{Experiments and Evaluation}
\input{contents/4-experiments}

\section{Limitations and Future Work}
\input{contents/5-limitations-and-future-work}

\section{Conclusion}
\input{contents/6-conclusion}


\section*{Acknowledgment}
This work was supported by Clemson University's Virtual Prototyping of Autonomy Enabled Ground Systems (VIPR-GS), under Cooperative Agreement W56HZV-21-2-0001 with the US Army DEVCOM Ground Vehicle Systems Center (GVSC).

\bibliographystyle{IEEEtran}
\bibliography{references}

\end{document}

%% file: contents/0-abstract.tex
\begin{abstract}

Accurate 3D semantic perception is critical for safe autonomous navigation. However, supervised LiDAR segmentation remains tied to closed taxonomies and to the cost of point-wise manual annotation. Open-vocabulary methods avoid that cost by projecting the output of 2D vision-language models onto LiDAR and distilling it into a 3D network. These methods rely almost exclusively on voxel-based sparse convolutions, and point transformers have so far been limited to indoor environments, where 3D data is dense and bounded. 

We present \textit{Lang3DSeg}, which establishes a point transformer as the backbone for annotation-free open-vocabulary segmentation of outdoor 3D LiDAR data, and is trained from scratch without geometric pre-training. This training paradigm necessitates addressing the inherent noise in 2D-to-3D label projections; specifically, naive projection often suffers from depth ambiguity, where points behind an object are erroneously assigned its semantic label. 
We therefore composite masks using an explicit class-priority rule and truncate each projected instance at the first gap in its depth distribution, correcting the projection error directly rather than averaging it over registered sequences.

\textit{Lang3DSeg} achieves \textit{52.8\% mIoU} on nuScenes validation and \textit{41.4\%} on SemanticKITTI, the highest among published annotation-free methods on both benchmarks. Every 3D semantic segmentation is on a single LiDAR sweep, and inference operates in real-time without running vision-language models.

\end{abstract}

\begin{keywords} 
Deep Learning for Visual Perception, 
3D Point Clouds, 
Autonomous 3D Recognition, 
Vision-Language Models, 
Open-Vocabulary Segmentation
\end{keywords} 

%% file: contents/1-introduction.tex
Accurate 3D semantic perception underpins autonomous navigation safety in unstructured environments. Supervised point transformers reach high benchmark accuracy, but they depend on dense point-wise annotation and are bound to the closed taxonomy they were trained on. Both properties limit a robot that must operate around rare obstacles, unfamiliar vehicle types, or terrain the annotation schema never
anticipated to mark or label.

Open-vocabulary distillation offers a way out. A 2D vision-language model labels camera images, those labels are projected onto the LiDAR sweep, and a 3D network learns from the result without human annotation. What has not changed across this literature is the 3D backbone: most methods in the setting, from CLIP2Scene~\cite{chen2023clip2scene} through Better Call SAL~\cite{ovsep2024better}, OV3D~\cite{jiang2024open}, and GGSD~\cite{wang2024open}, build on a voxel-based sparse convolutional network, which imposes a spatial quantization not present in the raw point measurements. Point transformers remove that constraint and have been adopted for open-vocabulary segmentation indoor environments~\cite{spaceformer2026}, where scenes are dense and bounded. Outdoor LiDAR is the opposite regime: sweeps are sparse, ring-structured, and span a hundred meters, and to our knowledge no point transformer has been trained for annotation-free open-vocabulary segmentation in it.

\begin{figure}[t]
  \centering
  \includegraphics[width=\columnwidth]{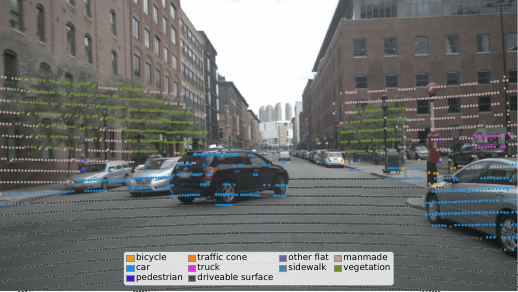}\\[1.5pt]
  \includegraphics[width=\columnwidth]{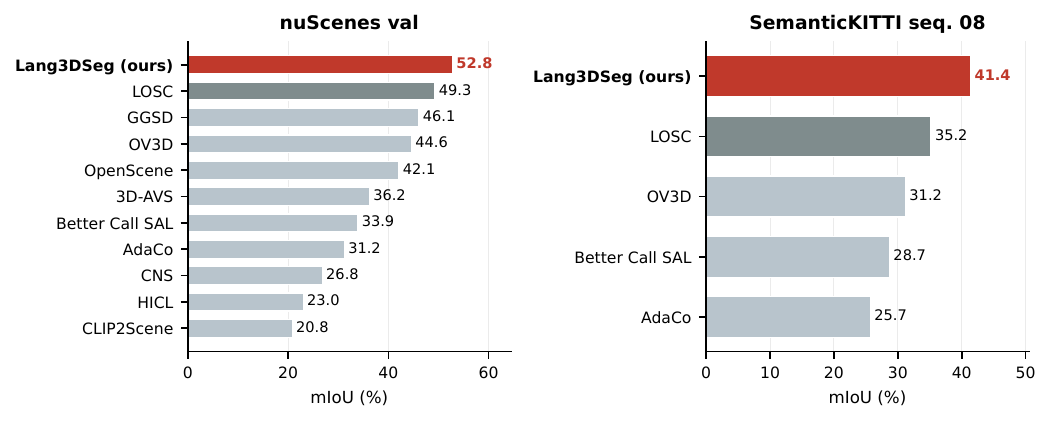}
  \caption{\textbf{Lang3DSeg} segments a single LiDAR sweep with no human
  annotation and no vision-language model at inference (top, predictions
  projected onto the front camera). A PointTransformerV3 backbone trained from
  scratch reports the highest mIoU among annotation-free methods on both
  nuScenes and SemanticKITTI (bottom).}
  \label{fig:teaser}
\end{figure}

\textbf{Lang3DSeg} closes that gap with two components. We train a PointTransformerV3~\cite{wu2024ptv3} backbone from scratch on open-vocabulary supervision, with no geometric pre-training and no supervised initialization; and because a from-scratch backbone has no prior to fall back on and reproduces the errors present in the projected pseudo-labels, we pair it with an offline curation stage that corrects those errors at the projection step rather than compensating for them afterwards. Together they outperform every published annotation-free result on both driving benchmarks.

Prior work treats the 2D-3D projection noise statistically, by temporal voting~\cite{samet2026losc}, adaptive label correction~\cite{zou2025adaco}, or keeping a vision-language model in the loop at inference~\cite{peng2023openscene}. Each method introduces computation or memory cost that may be unsuitable for resource-constrained deployment, and much of the error is not statistical in the first place: a point behind a car is mislabeled identically from every similar viewpoint, so voting reinforces it. We instead remove these labels at the projection stage itself. A per-pixel occlusion test does not suffice, since a point behind an object typically lands on its own unoccupied pixel inside that object's mask; we therefore test each mask instance as a whole for points that lie beyond the surface it describes.

The contributions of this paper are as follows.
\begin{itemize}\setlength{\itemsep}{1pt}
    \item We \textbf{create a new annotation-free open-vocabulary 3D learning framework}, and validate its effectiveness through 3D LiDAR outdoor semantic segmentation task by training a point transformer from scratch without projected pseudo-label errors.
    \item We \textbf{identify and quantify occlusion bleed} as a dominant and
    systematic error during 2D-to-3D label projection, concentrated in the
    foreground categories that significantly matter for class-averaged accuracy.
    \item We \textbf{introduce} a priority-ordered rasterization with an instance-wise depth approach to handle this error, requiring neither registration nor cross-frame voting.
    \item We \textbf{attain the SOTA annotation-free accuracy} on
    both benchmarks, 52.8\% mIoU on nuScenes and 41.4\% on SemanticKITTI, while
    keeping the vision-language model out of the inference path.
\end{itemize}

%% file: contents/2-related_work.tex
Open-vocabulary 3D scene understanding largely proceeds by transferring
knowledge from 2D vision-language models~\cite{zhang2024vision}. We organize the
relevant work by where the foundation model sits relative to inference, by the 3D
backbone each method adopts, and how each handles projection noise.

\subsection{Cross-Modal Distillation and Open-Vocabulary Frameworks}
Early zero-shot methods keep the 2D model in the loop at test time.
PointCLIP~\cite{zhang2022pointclip} and OpenScene~\cite{peng2023openscene}
project point clouds onto image planes and aggregate
CLIP~\cite{radford2021learning} features during inference, and
OpenMask3D~\cite{takmaz2023openmask3d} refines projected masks with
SAM~\cite{kirillov2023segment} before extracting free-form features. Accuracy is
competitive, but evaluating a vision-language model across several cameras for
every sweep is not viable on an embedded platform.

Offline distillation removes that cost. ULIP~\cite{xue2023ulip} aligns points to
text directly, while SLidR~\cite{sautier2022slidr}, Seal~\cite{liu2023segment} and
CLIP2Scene~\cite{chen2023clip2scene} learn correspondences across driving
sequences that can later be fine-tuned. Better Call SAL~\cite{ovsep2024better},
OV3D~\cite{jiang2024open}, GGSD~\cite{wang2024open} and AFOV~\cite{sun2025afov}
generate pseudo-labels offline and distill them into 3D-native networks, which is
the setting we adopt. AFOV additionally pre-trains the backbone with a tri-modal
contrastive objective, and LOSC~\cite{samet2026losc} initializes from
ScaLR~\cite{puy2024scalr}; both therefore depend on a self-supervised stage that
we do not use.

\subsection{Backbones for Open-Vocabulary LiDAR Segmentation}
What has not varied across this literature is the 3D architecture.
CLIP2Scene~\cite{chen2023clip2scene} and AdaCo~\cite{zou2025adaco} use Sparse Point-Voxel Convolution (SPVCNN)~\cite{tang2020searching};
CNS~\cite{chen2023towards}, OpenScene~\cite{peng2023openscene}, Better Call
SAL~\cite{ovsep2024better}, 3D-AVS~\cite{wei20253d}, OV3D~\cite{jiang2024open},
GGSD~\cite{wang2024open} and AFOV~\cite{sun2025afov} use MinkowskiNet~\cite{choy20194d} or
SparseConvNet; LOSC~\cite{samet2026losc} uses WaffleIron~\cite{puy23waffleiron}. Every one either
voxelizes the sweep or projects it onto a dense grid.

Point transformers instead attend over serialized neighborhoods of the raw
points, and PointTransformerV3~\cite{wu2024ptv3} scales to the point counts a
LiDAR sweep produces. The architecture has reached open-vocabulary segmentation
indoor environments: SpaCeFormer~\cite{spaceformer2026} couples a space-curve transformer
with sliding-window attention for instance segmentation and reports
PointTransformerV3 as a strong baseline there. Its evaluation is on ScanNet200
and Replica indoor scenes, where scenes are dense, bounded and roughly uniformly sampled;
outdoor LiDAR is the opposite regime, with point density falling off
quadratically with range. To our knowledge, Lang3DSeg is the first to distill
open-vocabulary supervision into a point transformer for outdoor LiDAR
segmentation, and the first to do so from scratch.

\subsection{Noise Mitigation and Geometric Consistency}
The treatment of projection noise can substantially affect segmentation accuracy. Better
Call SAL~\cite{ovsep2024better} applies DBSCAN~\cite{khan2014dbscan} within individual frames to
discard spatial outliers. LOSC~\cite{samet2026losc} consolidates labels across a
voxelized sequence by temporal voting, which suppresses per-frame flicker
effectively and currently gives the strongest published results in this setting.
AdaCo~\cite{zou2025adaco} corrects supervision during training through an
adaptive label corrector, and AFOV~\cite{sun2025afov} propagates labels through
an approximate-flat interaction module that fills unobserved regions.

These approaches primarily address projection noise as a stochastic quantity to be averaged or
down-weighted, and much of it is neither. The label assigned to a point behind a
car is wrong identically from every similar viewpoint, so temporal voting
reinforces it; and the quantity that would reject it, its range along the camera
ray, is available at projection time and discarded. Lang3DSeg filters at the
source instead, cutting each projected instance at the first gap in its depth
distribution and resolving the residual ground ambiguity with a per-sweep plane
model, as Sec.~\ref{sec:methods} details.

%% file: figures/framework.tex
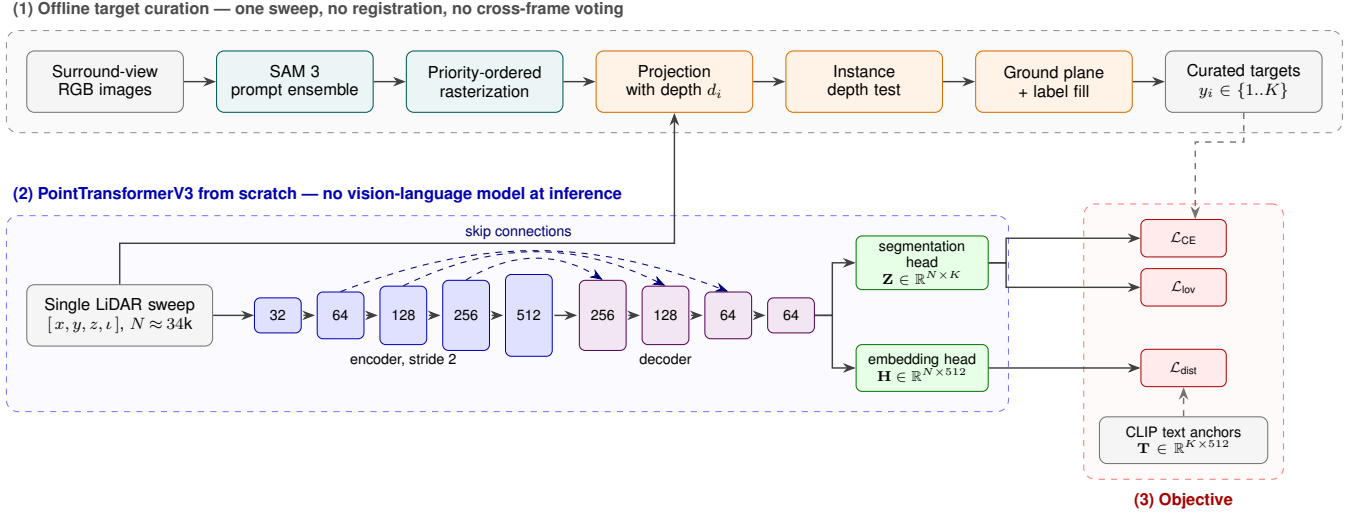
\begin{figure*}[t]
\vspace{6pt}
\centering
\resizebox{\textwidth}{!}{%
\begin{tikzpicture}[
    >={Stealth[length=2.2mm]}, font=\sffamily\small,
    node distance=5mm and 5.5mm,
    box/.style   = {draw, rounded corners=1.3mm, minimum height=10.5mm,
                    text width=24mm, align=center, inner sep=1.4mm, line width=0.5pt,
                    font=\sffamily\footnotesize,
                    execute at begin node={\hyphenpenalty=10000\relax}},
    twod/.style  = {box, draw=teal!70!black,   fill=teal!8},
    geo/.style   = {box, draw=orange!85!black, fill=orange!10},
    dat/.style   = {box, draw=black!55,        fill=black!4},
    enc/.style   = {draw=blue!70!black, fill=blue!12, rounded corners=1mm,
                    minimum width=8mm, align=center, inner sep=0.8mm,
                    font=\sffamily\scriptsize},
    dec/.style   = {draw=violet!65!black, fill=violet!10, rounded corners=1mm,
                    minimum width=8mm, align=center, inner sep=0.8mm,
                    font=\sffamily\scriptsize},
    hd/.style    = {draw=green!45!black, fill=green!10, rounded corners=1mm,
                    minimum height=8mm, text width=21mm, align=center,
                    inner sep=0.8mm, font=\sffamily\scriptsize},
    lss/.style   = {draw=red!65!black, fill=red!8, rounded corners=1mm,
                    minimum height=6.4mm, text width=13mm, align=center,
                    inner sep=0.8mm, font=\sffamily\scriptsize},
    fl/.style    = {->, line width=0.7pt, draw=black!75},
    sup/.style   = {->, line width=0.7pt, draw=black!55, dashed},
    skip/.style  = {->, line width=0.5pt, draw=blue!45!black, dashed},
    lbl/.style   = {font=\sffamily\scriptsize, inner sep=1pt},
    ttl/.style   = {font=\sffamily\bfseries\footnotesize}
]

\node[dat]                  (img)  {Surround-view\\RGB images};
\node[twod, right=of img]   (sam)  {SAM 3\\prompt ensemble};
\node[twod, right=of sam]   (pri)  {Priority-ordered\\rasterization};
\node[geo,  right=of pri]   (proj) {Projection\\with depth $d_i$};
\node[geo,  right=of proj]  (occ)  {Instance\\depth test};
\node[geo,  right=of occ]   (grd)  {Ground plane\\+ label fill};
\node[dat,  right=of grd]   (lab)  {Curated targets\\$y_i\!\in\!\{1..K\}$};
\foreach \a/\b in {img/sam, sam/pri, pri/proj, proj/occ, occ/grd, grd/lab}
  \draw[fl] (\a) -- (\b);

\node[dat, below=40mm of img.north west, anchor=north west, text width=29mm]
      (sweep) {Single LiDAR sweep\\$[\,x,y,z,\iota\,]$, $N\!\approx\!34$k};
\node[enc, right=7mm of sweep, minimum height=6mm]  (e1) {32};
\node[enc, right=2.6mm of e1,  minimum height=8mm]  (e2) {64};
\node[enc, right=2.6mm of e2,  minimum height=10mm] (e3) {128};
\node[enc, right=2.6mm of e3,  minimum height=12mm] (e4) {256};
\node[enc, right=2.6mm of e4,  minimum height=14mm] (e5) {512};
\node[dec, right=4.5mm of e5,  minimum height=12mm] (d4) {256};
\node[dec, right=2.6mm of d4,  minimum height=10mm] (d3) {128};
\node[dec, right=2.6mm of d3,  minimum height=8mm]  (d2) {64};
\node[dec, right=2.6mm of d2,  minimum height=6mm]  (d1) {64};
\draw[fl] (sweep) -- (e1);
\foreach \a/\b in {e1/e2,e2/e3,e3/e4,e4/e5,e5/d4,d4/d3,d3/d2,d2/d1}
  \draw[fl, shorten >=0.3mm, shorten <=0.3mm] (\a) -- (\b);
\draw[skip] (e4.north) to[bend left=32] (d4.north);
\draw[skip] (e3.north) to[bend left=26] (d3.north);
\draw[skip] (e2.north) to[bend left=21] (d2.north);
\node[lbl, text=blue!45!black, above=6.5mm of e4, xshift=9mm] {skip connections};
\node[lbl, below=1mm of e3] {encoder, stride 2};
\node[lbl, below=1mm of d3] {decoder};

\node[hd, right=7mm of d1, yshift=9mm]  (seg) {segmentation head\\$\mathbf{Z}\!\in\!\mathbb{R}^{N\times K}$};
\node[hd, right=7mm of d1, yshift=-9mm] (emb) {embedding head\\$\mathbf{H}\!\in\!\mathbb{R}^{N\times512}$};
\draw[fl] (d1.east) -- ++(3mm,0) |- (seg.west);
\draw[fl] (d1.east) -- ++(3mm,0) |- (emb.west);

\node[lss, right=26mm of seg, yshift=4.2mm] (ce)  {$\mathcal{L}_{\text{CE}}$};
\node[lss, right=26mm of seg, yshift=-4.2mm](lov) {$\mathcal{L}_{\text{lov}}$};
\node[lss, right=26mm of emb]               (dis) {$\mathcal{L}_{\text{dist}}$};
\node[dat, below=5mm of dis, text width=26mm, minimum height=8mm, font=\sffamily\scriptsize] (txt)
      {CLIP text anchors\\$\mathbf{T}\!\in\!\mathbb{R}^{K\times512}$};
\draw[fl] (seg.east) -- ++(3mm,0) |- (ce.west);
\draw[fl] (seg.east) -- ++(3mm,0) |- (lov.west);
\draw[fl] (emb.east) -- (dis.west);
\draw[sup] (txt.north) -- (dis.south);
\draw[sup] (lab.south) -- ++(0,-5mm) -| ($(ce.north)+(2mm,0)$);
\draw[fl]  (sweep.north) -- ++(0,7mm) -| (proj.south);

\begin{scope}[on background layer]
  \node[draw=black!45, dashed, rounded corners=2mm, fill=black!2, inner sep=3.4mm,
        fit=(img)(lab)] (g1) {};
  \node[draw=blue!55, dashed, rounded corners=2mm, fill=blue!2, inner sep=3.4mm,
        fit=(sweep)(e5)(d1)(seg)(emb)] (g2) {};
  \node[draw=red!55, dashed, rounded corners=2mm, fill=red!2, inner sep=2.6mm,
        fit=(ce)(lov)(dis)(txt)] (g3) {};
\end{scope}
\node[ttl, text=black!70,      anchor=south west] at ([yshift=1mm]g1.north west)
      {(1) Offline target curation \textemdash{} one sweep, no registration, no cross-frame voting};
\node[ttl, text=blue!70!black, anchor=south west] at ([yshift=1mm]g2.north west)
      {(2) PointTransformerV3 from scratch \textemdash{} no vision-language model at inference};
\node[ttl, text=red!65!black,  anchor=north]      at ([yshift=-1mm]g3.south)
      {(3) Objective};
\end{tikzpicture}
}
\caption{The Lang3DSeg framework. \textbf{(1)} Targets are curated offline.
SAM~3 is prompted with a descriptive ensemble, masks are rasterized under an
explicit class priority so that large background regions cannot overwrite small
foreground objects, and each projected instance is cut at the first gap in its
depth distribution $d_i$, discarding points that lie behind the mask that
claimed them. A per-sweep RANSAC ground model then separates flat terrain from
vertical vegetation, and unlabeled points are filled only where neighboring
labels agree. \textbf{(2)} A PointTransformerV3 backbone is trained from scratch
on these targets; numbers give the channel width of each encoder and decoder
stage. \textbf{(3)} Training combines weighted cross-entropy, Lov\'asz-Softmax,
and a cosine alignment term that draws the point embeddings toward frozen CLIP
text anchors. The anchors supervise training only and never enter the forward
path, so inference consumes one LiDAR sweep and no image data.}
\label{fig:architecture}
\end{figure*}

%% file: contents/3-methods.tex
\label{sec:methods}

The difficulty is not obtaining 2D masks but transferring them to LiDAR. A
backbone trained from scratch has no prior with which to discount projection
error, so we treat target curation as a first-class component and resolve that
error offline, using only the information in one sweep.
Fig.~\ref{fig:architecture} summarizes the three stages.

\subsection{Offline Target Curation}
\label{sec:curation}

\noindent\textbf{Descriptive prompting.} We query
SAM~3~\cite{carion2026sam} with a small ensemble of
natural-language phrases per class rather than the dataset's category string,
since driving taxonomies contain labels such as \textit{other flat} that no
foundation model recognizes. Each phrase is issued as an independent concept
query and its masks are collected separately. Prompt design is constrained by
the benchmark definition rather than by intuition: nuScenes assigns parking lots
to \textit{drivable surface} and plazas to \textit{sidewalk}, so prompting for
those under \textit{other flat} removes points from two well-performing classes
in exchange for an unreliable gain on a rare one. Our ensembles overlap with
those published by LOSC~\cite{samet2026losc} for the vegetation, terrain,
trailer and construction categories, which we adopt directly. Table~\ref{tab:prompts} lists every
phrase and the projection priority of each class. SAM~3 operates in video mode
over windows of five consecutive keyframes, which amortizes the cost of text
grounding; this is 2D mask tracking only, and no point clouds are registered or
merged at any stage.

\input{tables/prompt_table}

\noindent\textbf{Priority-ordered rasterization.} Masks from all classes must be
composited into a single 2D label image, and the compositing order matters more
than it appears. A drivable-surface mask covers most of the frame and carries a
high confidence score, so a confidence-ordered rule erases the traffic cones and
pedestrians standing on it. We instead assign each class $c$ a fixed priority
$\pi(c)$, with thin and small foreground categories highest and the road surface
lowest, and resolve every pixel by
\begin{equation}
\text{write} \iff
\big(\pi(c) > \pi^{\star}\big) \;\lor\;
\big(\pi(c) = \pi^{\star} \;\land\; s > s^{\star}\big),
\label{eq:priority}
\end{equation}
where $\pi^{\star}$ and $s^{\star}$ are the priority and score currently stored
at that pixel and $s$ is the SAM~3 mask score. Priority decides between classes;
confidence only breaks ties within a class.

\noindent\textbf{Projection and occlusion rejection.} Points are transformed
into each camera frame, and we retain the depth $d_i$ along the optical axis that
the usual projection discards. Every point landing inside a mask inherits its
class and instance identifier, so the set mixes the object with its background. A per-pixel depth buffer does not help, since the occluded point
projects to a \emph{different} pixel that is also inside the mask. We instead
exploit the fact that a physical object is contiguous in depth while the surface
behind it is separated by empty space. For each instance $m$ we sort the depths of its
points, $d_{(1)} \le \dots \le d_{(n)}$, and retain the prefix up to the first
sufficiently large gap,
\begin{equation}
k^{\star} = \min\{\,k : d_{(k+1)} - d_{(k)} > \tau_d\,\},
\mathcal{K}_m = \{(1),\dots,(k^{\star})\},
\label{eq:depthgap}
\end{equation}
keeping all points if no such gap exists. With $\tau_d = 2.5$\,m this removes a
wall several metres behind a car while leaving an extended object such as a bus
or trailer intact, since its own depth extent is continuous. The test is applied
to foreground classes only, as background regions are legitimately
depth-extended. Points visible to several cameras are merged with the rule of
Eq.~\ref{eq:priority}, so overlapping views act as a weak ensemble.

\noindent\textbf{Ground-plane purification.} 2D segmenters confuse flat terrain
with vertical vegetation, since a grass verge and a tree canopy can be
indistinguishable in appearance; we resolve this geometrically. A plane
$(\mathbf{n}, d)$ with $\Vert\mathbf{n}\Vert_2 = 1$ is fitted by RANSAC~\cite{schnabel2007efficient} to the
points labeled \textit{drivable surface} within 40\,m, rejecting hypotheses with
$n_z < 0.90$ and refining the consensus set by least squares. Each point then
has a signed elevation $e_i = \mathbf{n}^{\!\top} p_i + d$. A point labeled
\textit{vegetation} is reassigned to \textit{terrain} when it lies close to the
ground and is not part of a vertical structure,
\begin{equation}
|e_i| < \tau_e
\;\land\;
\max_{p_j \in \mathcal{N}_r(p_i)} z_j - \min_{p_j \in \mathcal{N}_r(p_i)} z_j < \tau_z ,
\label{eq:veg}
\end{equation}
with $\tau_e = 0.35$\,m and $\tau_z = 0.5$\,m over a radius-$r$ neighborhood,
$r = 1.0$\,m. The second condition protects tree trunks, whose bases sit near the
ground but whose neighborhoods span several metres vertically.

Within 15\,m we additionally require local planarity. From the covariance
$\mathbf{C}_i = |\mathcal{N}_r|^{-1}\sum_{p_j \in \mathcal{N}_r(p_i)}
(p_j - \bar{p})(p_j - \bar{p})^{\!\top}$
with eigenvalues $\lambda_1 \ge \lambda_2 \ge \lambda_3$ and eigenvectors
$\mathbf{v}_1,\mathbf{v}_2,\mathbf{v}_3$, we use planarity
$P_i = (\lambda_2 - \lambda_3)/\lambda_1$, surface variation
$v_i = \lambda_3 / \sum_k \lambda_k$ and the normal $\mathbf{v}_3$, requiring
$P_i > 0.35$, $v_i < 0.20$ and $|v_{3,z}| > 0.80$.

The near-field restriction follows from the sampling. On a 32-beam sensor at
height $h$, a beam at depression angle $\theta$ meets the ground at
$r = h/\tan\theta$, so consecutive rings are separated by
$\Delta r \approx h\,\Delta\theta/\sin^2\theta$: about 1.3\,m at 10\,m range for
$h = 1.84$\,m and $\Delta\theta \approx 1.33^\circ$, but roughly 5\,m at 20\,m.
Beyond that a ground point's neighbors all lie along one ring whatever radius is
chosen, $\mathbf{C}_i$ degenerates toward rank one, and both $P_i$ and
$\mathbf{v}_3$ stop carrying information. The range-independent elevation test of
Eq.~\ref{eq:veg} therefore remains the primary signal.

\noindent\textbf{Label completion.} Camera coverage is incomplete, so part of
each sweep receives no label. We fill an unlabeled point from its $k$ nearest
labeled neighbors within 0.4\,m, and only when at least three are found and all
agree. The step is strictly additive, never overwriting an existing label, which
raises coverage without the churn unconstrained smoothing introduces.

\subsection{Network}
\label{sec:architecture}

\noindent\textbf{Backbone.} We use PointTransformerV3~\cite{wu2024ptv3}
initialized from scratch, with no geometric pre-training and no supervised
initialization. Encoder channel widths are $(32,64,128,256,512)$ with depths
$(2,2,2,6,2)$ and stride $2$ between stages; the decoder widths are
$(64,64,128,256)$, and points are serialized on a 0.05\,m grid using $z$-order
and Hilbert curves. The network consumes one sweep and predicts dense logits
$\mathbf{Z} \in \mathbb{R}^{N \times K}$ and embeddings
$\mathbf{H} \in \mathbb{R}^{N \times 512}$.

\noindent\textbf{Input representation.} Coordinates are centered in $x$ and $y$
only; the vertical axis is left in the sensor frame, because height above the
ground separates the very classes this work targets and because mean-centering
$z$ introduces a train--test shift whenever the training sample is drawn
differently from the full sweep used at inference. The per-point feature is
$[\,x', y', z, \iota\,]$, with $\iota$ the normalized return intensity.

\subsection{Objective}
\label{sec:optimization}

\noindent\textbf{Semantic terms.} We combine weighted cross-entropy with
Lov\'asz-Softmax~\cite{berman2018lovasz}, which optimizes the IoU surrogate
directly:
\begin{equation}
\mathcal{L}_{\text{CE}} = -\frac{1}{|\mathcal{V}|} \sum_{i \in \mathcal{V}} w_{y_i}
\log \frac{\exp(z_{i,y_i})}{\sum_{j=1}^{K}\exp(z_{i,j})},
\end{equation}
over the set $\mathcal{V}$ of points carrying a curated label. Class weights use
inverse square-root frequency, $w_c \propto f_c^{-1/2}$, normalized to unit mean
and clipped to $[0.2, 8]$, with $f_c$ measured on the curated targets
themselves. Inverse frequency without the square root over-weights rare classes enough to
destabilize training when combined with Lov\'asz-Softmax, which is itself warmed
up linearly over three epochs because it is poorly conditioned while the logits
are near-uniform.

\noindent\textbf{Cosine alignment.} A light alignment term places the point
embeddings in the CLIP text space,
\begin{equation}
\mathcal{L}_{\text{dist}} = \frac{1}{|\mathcal{V}|} \sum_{i \in \mathcal{V}}
\left(1 - \left\langle \frac{\mathbf{h}_i}{\Vert \mathbf{h}_i \Vert_2},\; \mathbf{t}_{y_i} \right\rangle \right),
\end{equation}
with $\mathbf{t}_{y_i}$ the unit-norm anchor of the assigned class, taken from
the frozen \texttt{clip-vit-base-patch32} text encoder after its projection into
the joint image--text space. The anchors enter training only through this term
and are absent from the forward path, so inference requires no text encoder.
Beyond regularization the term makes the embeddings queryable by text: a class
outside the training taxonomy can be scored by cosine similarity between
$\mathbf{H}$ and any text embedding, without retraining.

\noindent\textbf{Total.} The objective is
\begin{equation}
\mathcal{L} = \lambda_{\text{ce}}\mathcal{L}_{\text{CE}}
            + \lambda_{\text{lov}}\mathcal{L}_{\text{lov}}
            + \lambda_{\text{dist}}\mathcal{L}_{\text{dist}},
\end{equation}
with $\lambda_{\text{ce}} = \lambda_{\text{lov}} = 1.0$ and
$\lambda_{\text{dist}} = 0.1$, determined empirically.

%% file: tables/prompt_table.tex
\begin{table}[t]
\vspace{9pt}
\caption{Prompt ensembles issued to SAM~3 and the projection priority of each
class. Higher priority wins during rasterization; ties within a priority level
are broken by mask score. The ordering follows one rule: a category that occupies
few pixels and is easily overwritten sits above one that covers large contiguous
regions, and the road surface is the base layer.}
\label{tab:prompts}
\centering
\footnotesize
\setlength{\tabcolsep}{4pt}

\begin{tabular}{@{}llc@{}}
\toprule
\textbf{Class} & \textbf{Prompts} & \textbf{Priority} \\
\midrule

\rowcolor{gray!10}
traffic cone     & traffic cone; orange construction cone & 160 \\

pedestrian       & pedestrian; person & 150 \\

\rowcolor{gray!10}
bicycle          & bicycle & 145 \\

motorcycle       & motorcycle & 140 \\

\rowcolor{gray!10}
barrier          & barrier; barricade; guard rail & 130 \\

constr.\ vehicle & excavator; bulldozer; crane; & 125 \\
                 & concrete mixer truck & \\

\rowcolor{gray!10}
trailer          & semi trailer; cargo container; & 120 \\
\rowcolor{gray!10}
                 & shipping container & \\

truck            & truck & 115 \\

\rowcolor{gray!10}
bus              & bus & 110 \\

car              & car & 105 \\

\rowcolor{gray!10}
manmade          & building; wall; pole; fence & 60 \\

vegetation       & tree; bush; shrub & 55 \\

\rowcolor{gray!10}
other flat       & traffic island; raised median strip; & 45 \\
\rowcolor{gray!10}
                 & railroad tracks & \\

sidewalk         & sidewalk & 40 \\

\rowcolor{gray!10}
terrain          & grass; grassland; lawn; terrain & 35 \\

drivable surface & road; asphalt & 10 \\

\bottomrule
\end{tabular}
\end{table}


%% file: contents/4-experiments.tex
\label{sec:experiments}

\begin{figure*}[t]
\centering
\includegraphics[width=\linewidth]{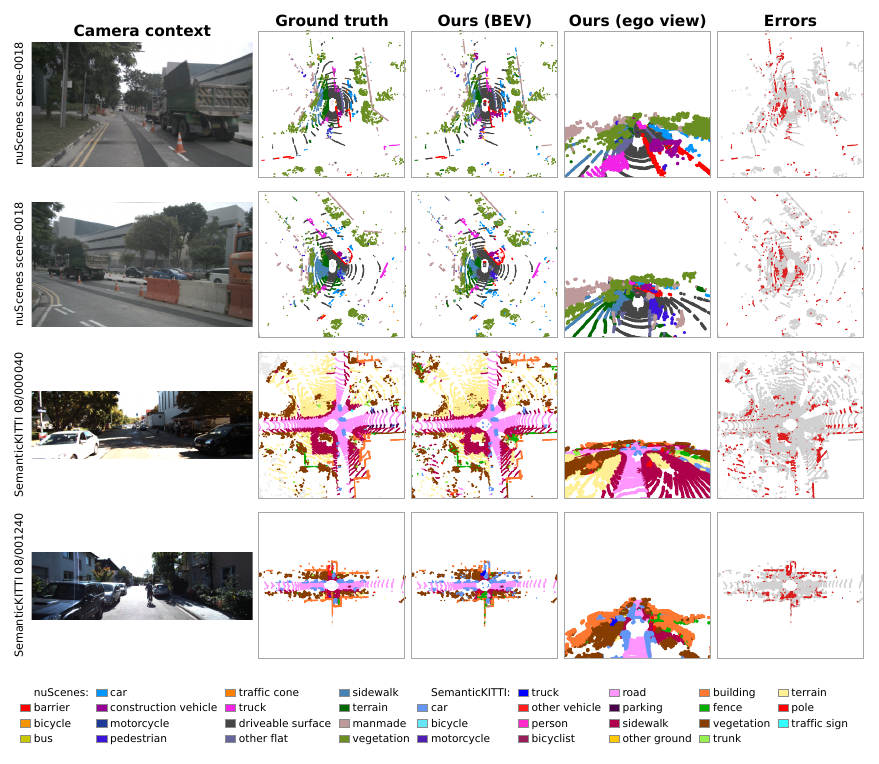}
\caption{Qualitative results on nuScenes validation (top two rows) and
SemanticKITTI sequence 08 (bottom two rows). Left to right: camera context;
ground-truth annotations; the Lang3DSeg prediction in bird's-eye view; the same
prediction in an ego-centric perspective view; and an error map with
misclassified points in red, correctly classified points in grey and unannotated
points in white. Lang3DSeg is trained without any human annotation and consumes
no image data at inference.}
\label{fig:qualitative}
\end{figure*}

\subsection{Experimental Setup}

\noindent\textbf{Datasets.} \textit{nuScenes}~\cite{caesar2020nuscenes} contains
1{,}000 scenes recorded with a 32-beam LiDAR and six surround-view cameras. We
curate targets on the 700-scene training split and evaluate on all 
frames of the 150-scene validation split over 16 categories.
\textit{SemanticKITTI}~\cite{behley2019semantickitti} provides 64-beam sequences
with a single forward camera; we use sequences 00-07, 09 and 10 for training and
sequence 08 for validation over 19 categories. No human annotation is used at any
point in curation or training; ground truth is used only to compute the reported
metrics.

\noindent\textbf{Metrics.} We report mean Intersection-over-Union over all
classes. Points labeled as ignore are excluded, and on SemanticKITTI the moving
categories are mapped back to their static counterparts, following the
single-scan benchmark protocol.

\noindent\textbf{Implementation.} Training uses AdamW with peak learning rate
$1\times10^{-4}$, weight decay $5\times10^{-3}$, a one-cycle schedule with 5\%
warmup and gradient clipping at 1.0, in full precision; mixed precision
interacts badly with the fused attention kernel, since a single overflow enters
the normalization statistics during the forward pass and cannot be undone by
skipping the optimizer step. We train on full unsampled sweeps, with an effective
batch of 12 on nuScenes and 16 on SemanticKITTI, augmenting by random rotation
about the vertical axis, random $x$ and $y$ mirroring, anisotropic scaling in
$[0.9,1.1]$, and per-point jitter with $\sigma = 5$\,mm. At inference we average
softmax outputs over four rotations about the vertical axis; latency is measured
without this averaging.

\begin{table*}[t]
\vspace{9pt}
\caption{Annotation-free 3D semantic segmentation on nuScenes validation and
SemanticKITTI sequence 08. \textit{VLM-free} indicates that no vision-language
model is evaluated at test time. \textit{Pre-train} indicates a self-supervised
initialization of the 3D backbone. Every prior method uses a voxel-based or
projection-based backbone.}
\label{tab:main_results}
\centering
\footnotesize
\setlength{\tabcolsep}{6pt}

\begin{tabular}{@{}llccllcc@{}}
\toprule
\textbf{Method} & \textbf{Venue} & \textbf{VLM-free} & \textbf{Pre-train} &
\textbf{3D backbone} & \textbf{Backbone type} & \textbf{nuScenes} &
\textbf{SemanticKITTI} \\
\midrule

\rowcolor{gray!18}
\multicolumn{8}{@{}l}{\textbf{Fully supervised upper bound} \hspace{2pt}
\textit{(trained from scratch)}} \\

PTv3~\cite{wu2024ptv3}
& CVPR'24 & -- & -- & PTv3 & point transformer & 80.4 & 70.8 \\

\midrule

\rowcolor{gray!18}
\multicolumn{8}{@{}l}{\textbf{Annotation-free methods}} \\

CLIP2Scene~\cite{chen2023clip2scene}
& CVPR'23 & \checkmark & -- & SPVCNN & sparse voxel & 20.8 & -- \\

\rowcolor{gray!8}
HICL~\cite{kang2024hierarchical}
& CVPR'24 & \checkmark & -- & SPVCNN & sparse voxel & 23.0 & -- \\

CNS~\cite{chen2023towards}
& NeurIPS'23 & \checkmark & -- & MinkowskiNet & sparse voxel & 26.8 & -- \\

\rowcolor{gray!8}
AdaCo~\cite{zou2025adaco}
& AAAI'25 & \checkmark & -- & SPVCNN & sparse voxel & 31.2 & 25.7 \\

Better Call SAL~\cite{ovsep2024better}
& ECCV'24 & \checkmark & -- & MinkowskiNet & sparse voxel & 33.9 & 28.7 \\

\rowcolor{gray!8}
3D-AVS~\cite{wei20253d}
& CVPR'25 & \checkmark & -- & MinkowskiNet & sparse voxel & 36.2 & -- \\

OpenScene~\cite{peng2023openscene}
& CVPR'23 & -- & -- & MinkowskiNet & sparse voxel & 42.1 & -- \\

\rowcolor{gray!8}
OV3D~\cite{jiang2024open}
& CVPR'24 & \checkmark & -- & MinkowskiNet & sparse voxel & 44.6 & 31.2 \\

GGSD~\cite{wang2024open}
& ECCV'24 & \checkmark & -- & MinkowskiNet & sparse voxel & 46.1 & -- \\

\rowcolor{gray!8}
AFOV~\cite{sun2025afov}
& AAAI'25 & \checkmark & \checkmark & MinkowskiNet & sparse voxel & 47.7 & -- \\

LOSC~\cite{samet2026losc}
& 3DV'26 & \checkmark & \checkmark & WaffleIron & projection & 49.3 & 35.2 \\

\midrule

\rowcolor{green!10}
\textbf{Lang3DSeg (ours)}
& & \checkmark & -- & PTv3 &
\textbf{point transformer} & \textbf{52.8} & \textbf{41.4} \\

\bottomrule
\end{tabular}
\end{table*}

\subsection{Benchmark Comparison}

Table~\ref{tab:main_results} places Lang3DSeg against published annotation-free
results. On nuScenes validation it reaches \textbf{52.8\% mIoU} and on
SemanticKITTI sequence 08 \textbf{41.4\%}, the highest reported figures in this
setting on both benchmarks. The comparison is favorable on cost as well as
accuracy. LOSC reaches 49.3\% by consolidating labels across registered
sequences, initializing from a self-supervised checkpoint~\cite{puy2024scalr},
and retraining iteratively; AFOV reaches 47.7\% with a tri-modal contrastive
pre-training stage; OpenScene reaches 42.1\% but evaluates a vision-language
model across six cameras for every sweep. Consistent with existing counterparts, we did not benchmark our method against UniPLV~\cite{wang2024uniplv} as it is not peer-reviewed. Table~\ref{tab:perclass} gives the per-class breakdown against LOSC, the strongest published baseline.

Lang3DSeg uses a single training run, no self-supervised initialization, and no image data at test time.
The backbone column is worth reading alongside the accuracy. Every prior method
in the table voxelizes the sweep or projects it onto a dense grid. Lang3DSeg is
the only entry built on a point transformer, and the only one trained from
scratch without a pre-training stage. We do not over-attribute the margin. We ran no controlled backbone substitution,
so we cannot separate the contribution of the architecture from that of the
curated targets. What we can measure is the effect of the curation, which
Table~\ref{tab:ablation_labels} isolates.

\noindent\textbf{SemanticKITTI.} A single forward-facing camera supervises only
the frontal frustum, so most of each sweep carries no 2D label during training.
Restricting evaluation to the frustum gives \textbf{47.8\%} mIoU against
\textbf{41.4\%} over the full sweep; the gap measures how far the backbone
extrapolates learned geometry into regions it never saw supervised.

\noindent\textbf{Runtime.} On a single NVIDIA H200, Lang3DSeg processes a
complete 360$^\circ$ nuScenes sweep in about \textbf{52.6\,ms}, over 19 frames per
second at batch size one without test-time augmentation and comfortably above
the 10\,Hz rotation rate of the sensor; the denser SemanticKITTI sweeps take
about 88\,ms. Methods that query a vision-language model at inference must
additionally run it over every camera.

\subsection{Ablation Studies}

\input{tables/perclass_table}

We ablate the curation pipeline directly on the labels it produces, scoring them
against ground truth without training anything. Label quality can be evaluated in
minutes rather than GPU-days, and it isolates the stage under study from the
variance of a training run.

\begin{table}[t]
\vspace{9pt}
\caption{Quality of the curated targets, measured against ground truth on 1{,}586
validation frames without any training. \textit{Labeled} scores only points that
carry a pseudo-label; \textit{all} charges unlabeled points as errors and
therefore reflects coverage as well as correctness.}
\label{tab:ablation_labels}
\centering
\footnotesize
\setlength{\tabcolsep}{5pt}

\begin{tabular}{@{}lccc@{}}
\toprule
\textbf{Curation stage} & \textbf{mIoU, labeled} & \textbf{mIoU, all} & \textbf{Coverage} \\
\midrule

\rowcolor{gray!10}
Projection with class priority  & 56.4 & 40.6 & 46.1 \\

\quad + occlusion depth test    & \textbf{58.7} & 41.1 & 45.5 \\

\rowcolor{gray!10}
\quad + ground-plane refinement & 58.6 & 41.1 & 45.5 \\

\quad + label completion        & 57.1 & \textbf{44.4} & \textbf{49.2} \\

\bottomrule
\end{tabular}
\end{table}

\noindent\textbf{Target curation.} Table~\ref{tab:ablation_labels} isolates each
stage. The occlusion depth test is the clear contributor, raising the quality of
the labels it keeps by 2.3 points, and \textit{manmade} absorbs most of the error
it removes because walls and buildings frequently form background surfaces behind foreground objects. Per class
it lifts \textit{car} by 9.1 points, \textit{pedestrian} by 7.2 and
\textit{barrier} by 6.1, but is not uniformly beneficial: \textit{trailer} loses
5.9, because a trailer is long enough that part of it occasionally falls beyond
the depth gap.

Ground-plane refinement is neutral on this metric, moving labeled mIoU by $-0.1$:
it corrects a visually obvious failure, terrain labeled as vegetation, but the
points involved are too few to register in a class-averaged score. Label
completion trades 1.6 points of per-point accuracy for 3.7 points of coverage,
gaining 3.3 on the metric that charges unlabeled points as errors. That is the
correct trade for a student network that must predict every point, and the reason we
report both columns.

\noindent\textbf{Negative results.} We also evaluated three additional design choices that did not improve performance but may be reasonable alternatives for future implementations. Verifying every mask by
comparing a CLIP crop embedding against the class ensembles rejected more than
half of the correct foreground detections, because the ensemble mean for a
visually diverse class such as \textit{manmade} sits near the centroid of
street-scene text embeddings. Concatenating per-mask CLIP features to the network
input also degraded accuracy: the feature vector is null exactly where no
pseudo-label exists, supplying a shortcut correlated with the loss mask rather
than the geometry. Routing the text anchors into the backbone through a cross-attention branch,
gated or otherwise, cost 1.6 to 1.9 mIoU against leaving the forward path purely
geometric, which is why the network we report is an unmodified
PointTransformerV3: the alignment term already supplies the benefit provided by the anchors, and the branch adds parameters a from-scratch backbone must fit from
noisy supervision. Finally, initializing from a self-supervised checkpoint
trained predominantly on indoor scans gave no improvement over training from
scratch, in contrast with the gain LOSC reports from an initialization learned on
driving data~\cite{puy2024scalr}, suggesting the benefit of pre-training here is
domain-specific rather than generic.

\subsection{Qualitative Results}

Fig.~\ref{fig:qualitative} shows predictions on both benchmarks against ground
truth, in bird's-eye and ego-centric views, with an error map isolating
misclassified points. Two patterns are visible. Boundaries between continuous
ground surfaces and vertical structure are clean, which is what the ground-plane
refinement is for even though it does not move the aggregate metric. Errors
concentrate on small and thin objects at range and along the boundaries between
co-planar ground classes, where neither geometry nor a single sweep provides
enough evidence to separate them; we return to both in Sec.~\ref{sec:limitations}.

%% file: tables/perclass_table.tex
\begin{table}[t]
\caption{Per-class IoU (\%) on nuScenes validation. LOSC is the strongest
published annotation-free baseline; $\Delta$ is our margin over it.}
\label{tab:perclass}
\centering
\footnotesize
\setlength{\tabcolsep}{5pt}
\begin{tabular}{@{}lccr@{}}
\toprule
\textbf{Class} & \textbf{LOSC}~\cite{samet2026losc} & \textbf{Lang3DSeg} & $\Delta$ \\
\midrule
\rowcolor{gray!8}
barrier              & 5.2  & \textbf{50.2} & $+45.0$ \\
bicycle              & \textbf{21.3} & 8.8  & $-12.5$ \\
\rowcolor{gray!8}
bus                  & \textbf{80.5} & 69.0 & $-11.5$ \\
car                  & 67.5 & \textbf{73.9} & $+6.4$ \\
\rowcolor{gray!8}
construction vehicle & 10.6 & \textbf{27.6} & $+17.0$ \\
motorcycle           & \textbf{64.1} & 46.4 & $-17.7$ \\
\rowcolor{gray!8}
pedestrian           & \textbf{72.7} & 64.5 & $-8.2$ \\
traffic cone         & \textbf{46.3} & 27.5 & $-18.8$ \\
\rowcolor{gray!8}
trailer              & 0.0  & \textbf{25.9} & $+25.9$ \\
truck                & 58.2 & \textbf{67.0} & $+8.8$ \\
\rowcolor{gray!8}
drivable surface     & 90.1 & \textbf{90.4} & $+0.3$ \\
other flat           & 0.0  & \textbf{23.3} & $+23.3$ \\
\rowcolor{gray!8}
sidewalk             & 51.0 & \textbf{55.9} & $+4.9$ \\
terrain              & \textbf{61.5} & 54.1 & $-7.4$ \\
\rowcolor{gray!8}
manmade              & 77.6 & \textbf{78.6} & $+1.0$ \\
vegetation           & 81.7 & \textbf{81.8} & $+0.1$ \\
\midrule
\textbf{mIoU}        & 49.3 & \textbf{52.8} & $+3.5$ \\
\bottomrule
\end{tabular}
\end{table}

%% file: contents/5-limitations-and-future-work.tex
\label{sec:limitations}

\noindent\textbf{Extended objects defeat the depth test.} The occlusion test
assumes an object is contiguous in depth while the surface behind it is separated
by empty space. That holds for a car in front of a wall and fails when an
object's own extent exceeds the threshold, which is why
Table~\ref{tab:ablation_labels} shows \textit{trailer} losing 5.9 IoU where
\textit{car} gains 9.1: a sparse return pattern along a long vehicle can itself
contain a gap wider than $\tau_d$. One threshold cannot serve a traffic cone and
an articulated trailer equally, and making $\tau_d$ class-dependent would
reintroduce the circularity the curation stage exists to avoid.

\noindent\textbf{Sparse and unobserved regions.} Two limits follow from using one
sweep and the cameras that saw it. Per-class accuracy trails methods that
consolidate across a registered sequence on \textit{motorcycle}, \textit{traffic
cone} and \textit{pedestrian}, because a pedestrian at 30\,m returns too few
points in a single sweep to separate from the structure behind it; consolidation
accumulates many views and gains real signal from doing so. And on SemanticKITTI
a single forward camera supervises only the frontal frustum, which the 47.8\%
against 41.4\% gap of Sec.~\ref{sec:experiments} quantifies. The backbone
extrapolates into unobserved regions, but on geometric regularity rather than
linguistic grounding, so novel objects in the blind spots remain out of reach.

\noindent\textbf{Co-planar ambiguity.} The ground model separates vertical
structure from flat ground because those differ in elevation, but cannot separate
two surfaces sharing a plane. An \textit{other flat} concrete island offers no
differential geometric cue against the \textit{drivable surface} beside it, and
the decision falls back on the 2D embedding, which is sensitive to lighting and
shadow. Our elevation banding recovers raised islands but not flush ones.

\noindent\textbf{Curation cost and closed-set inference.} Decoupling the
foundation models from inference moves the cost rather than removing it: curating
nuScenes takes on the order of 270 NVIDIA H200 GPU-hours before a single training step. Even though this curation cost is paid only once and never at deployment, it rules out adapting the vocabulary on-board. The deployed network also emits logits over a fixed set of classes; the
alignment term places the embeddings in the CLIP text space, so a novel category
could in principle be scored against any text embedding, but we have not
evaluated that and make no claim about it. 

\noindent\textbf{Experimental scope.} Controlled backbone substitution and margin divisions between the point transformer and the curated targets in Table~\ref{tab:main_results} are considered out of scope for this study.  
The variants in Sec.~\ref{sec:experiments} use a single seed each, so differences of about a point are not meaningful. The prompt ensembles are written by hand, as is standard here~\cite{samet2026losc}, and must be rewritten for a new taxonomy.

\noindent\textbf{Future work.} Making $\tau_d$ adaptive to each mask's spatial
extent would address the trailer case without class dependence; synthesizing
prompt ensembles with a language model would remove the manual step; and
relaxing the single-sweep constraint selectively, for sparse instances only,
would recover much of what consolidation offers at a fraction of its cost.

%% file: contents/6-conclusion.tex
\label{sec:conclusion}

We present Lang3DSeg, an annotation-free open-vocabulary framework for 3D
semantic segmentation built on a point transformer. We show that a
PointTransformerV3 backbone trains from scratch on open-vocabulary supervision
for outdoor LiDAR and sets a new bar on both driving benchmarks. Lang3DSeg achieves the state-of-the-art performance 52.8\% mIoU on nuScenes validation and 41.4\% on
SemanticKITTI sequence 08, without self-supervised initialization,
sequence-level consolidation or iterative retraining. Every 3D operation acts on
a single sweep, and the deployed network consumes only coordinates and intensity,
processing a full 360$^\circ$ scan in about 50\,ms. By keeping the visual-language foundation models entirely offline, the framework offers a practical basis for open-world perception for real-time 3D semantic segmentation.